\def\year{2021}\relax
\documentclass[letterpaper]{article} 
\usepackage{aaai21}  
\usepackage{times}  
\usepackage{helvet} 
\usepackage{courier}  
\usepackage[hyphens]{url}  
\usepackage{graphicx} 
\usepackage{natbib}  
\usepackage{caption} 
\usepackage{xcolor}
\usepackage{soul}
\usepackage{fdsymbol}
\usepackage{multirow}
\usepackage{multicol}
\usepackage{colortbl}
\usepackage{amsmath}

\DeclareMathOperator*{\argmax}{arg\,max}
\DeclareMathOperator*{\argmin}{arg\,min}

\definecolor{purple}{RGB}{210, 0, 210} 
\definecolor{green}{RGB}{0, 153, 0} 
\definecolor{grey1}{RGB}{255, 255, 255}
\definecolor{grey2}{RGB}{160, 160, 160}
\definecolor{grey3}{RGB}{0, 0, 0}

\usepackage{array}
\newcolumntype{L}[1]{>{\raggedright\let\newline\\\arraybackslash\hspace{0pt}}m{#1}}
\newcolumntype{C}[1]{>{\centering\let\newline\\\arraybackslash\hspace{0pt}}m{#1}}
\newcolumntype{R}[1]{>{\raggedleft\let\newline\\\arraybackslash\hspace{0pt}}m{#1}}

\title{Ergonomically Intelligent Physical Human-Robot Interaction:\\ Postural Estimation, Assessment, and Optimization\thanks{Research reported in this publication was supported in part by DARPA under grant N66001-19-2-4035 and the National Institute for Occupational Safety and Health under award number T420H008414-10.}}
\author{
    Amir Yazdani,\textsuperscript{\rm 1}Roya Sabbagh Novin,\textsuperscript{\rm 1\thanks{Roya Sabbagh Novin is currently with Embark Trucks Inc.}} Andrew Merryweather,\textsuperscript{\rm 1} Tucker Hermans\textsuperscript{\rm 1,}\textsuperscript{\rm 2}
    \\
}
\affiliations{
    \textsuperscript{\rm 1}University of Utah Robotics Center, SLC, UT; 
    \textsuperscript{\rm 2}NVIDIA Corporation, Seattle, WA;\\
    amir.yazdani@utah.edu
}

\begin{document}

\maketitle

\begin{abstract}
Ergonomics and human comfort are essential concerns in physical human-robot interaction. Common practical methods in the area either fail in estimating the correct posture due to occlusion or suffer from inaccurate ergonomics models in performing postural optimization. We propose a novel alternative framework for posture estimation, assessment, and optimization for ergonomically intelligent physical human-robot interaction. We show that we can estimate human posture solely from the trajectory of the interacting robot with median deviation of 5 deg from motion capture. We propose DULA, a differentiable ergonomics assessment tool with 99.73\% accuracy comparing to RULA. We use DULA in postural optimization for physical human-robot interaction tasks such as co-manipulation and teleoperation. We evaluate our framework through human and simulation experiments.
\end{abstract}

\section{Introduction}
\begin{figure}[t]
\center
\includegraphics[width = 8.5cm]{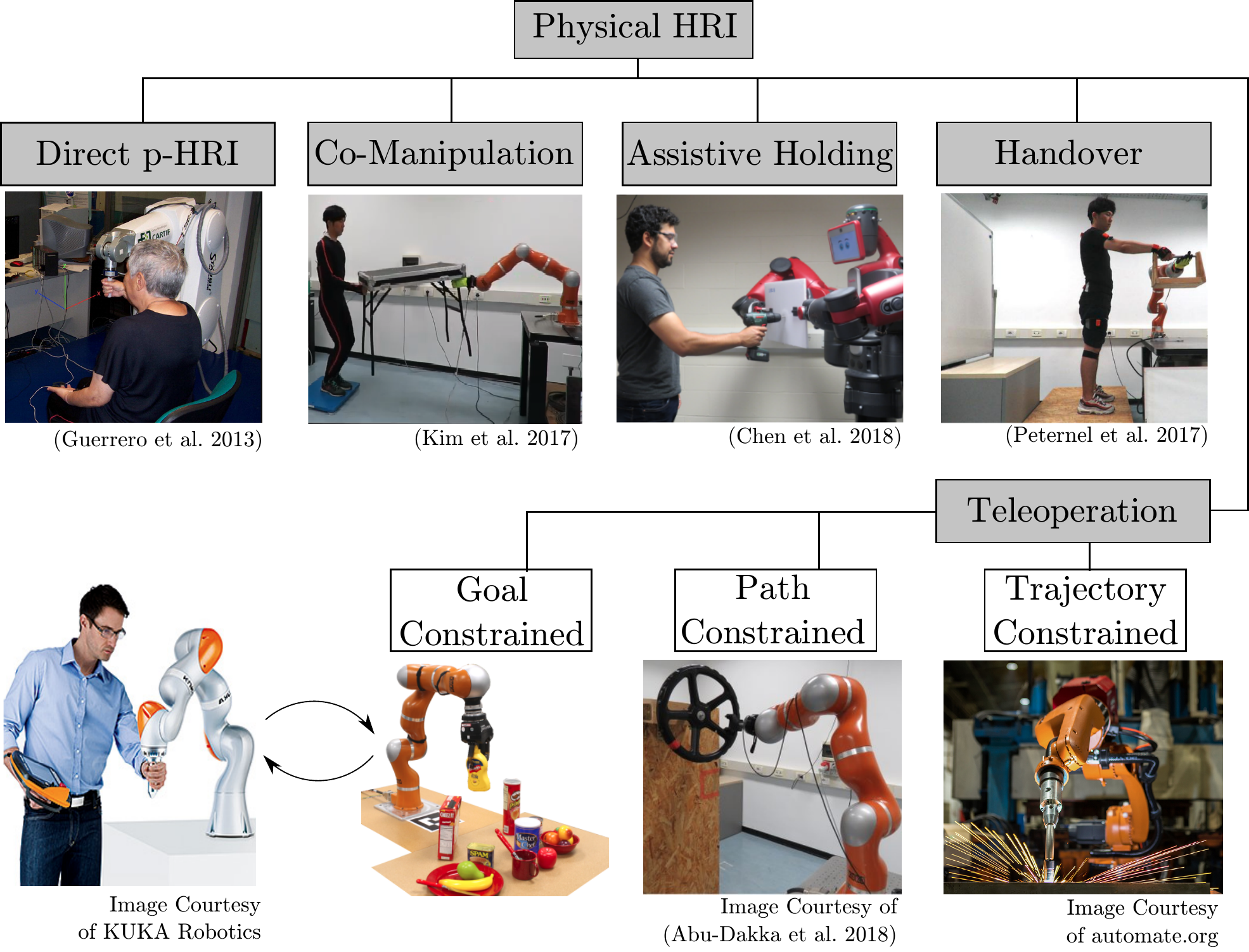}
\caption{\small Categories of physical HRI applications.}
\label{fig:p-HRI_taxonomy}
\end{figure}

Improving workplace ergonomics and reducing the risk of work-related musculoskeletal disorders~(WMSDs) has been the focus of researchers for many years~\cite{punnett2004work,erdinc2008ergonomics}. High rate of injuries in industry~\cite{schneider2001musculoskeletal,yu2012work} has further motivated these studies. WMSDs are the second-largest source of disabilities worldwide~\cite{vos2015global}, and awkward postures are the main contributor to the work-related injuries~\cite {da2010risk}.
\begin{figure}[t]
\center
\includegraphics[width = 7.9cm]{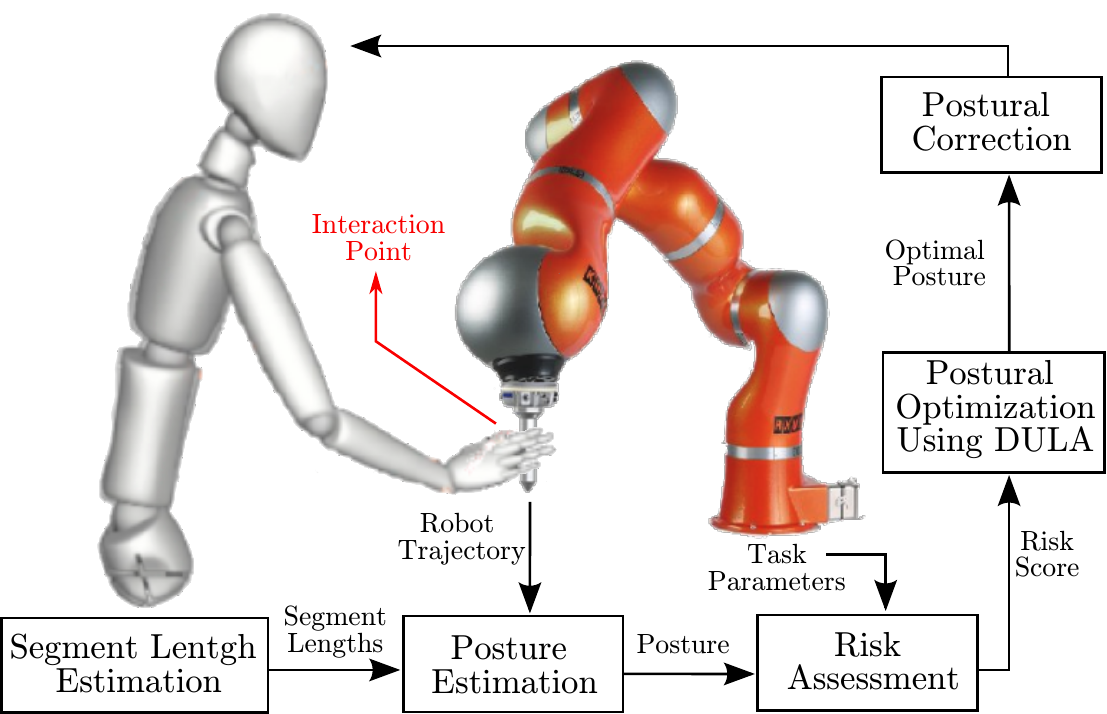}
\caption{\small The framework for ergonomically intelligent p-HRI.}
\label{fig:framework}
\end{figure}
Industry 4.0 initiative~\cite{gorecky2014human} and the development of collaborative workplaces where smart agents~(e.g., collaborative robots~\cite{sherwani2020collaborative}) physically collaborate with humans to perform different types of tasks, highlights the importance of human comfort and ergonomics in physical human-robot interaction~(p-HRI). In this paper, we introduce a framework for ergonomically intelligent p-HRI including posture estimation, assessment, optimization and correction. Figure \ref{fig:framework} shows our proposed framework. 

p-HRI covers a vast number of applications such as generic p-HRI which includes direct physical interaction with the robot (e.g., robotics physical therapy~\cite{guerrero2013using}), co-manipulation~\cite{kim2017anticipatory}, assistive holding~\cite{chen2018planning}, handover~\cite{busch2018planning}, and teleoperation~\cite{rahal2020caring}. We categorize teleoperation into three main types: \textit{goal-constrained teleoperation}~(e.g., pick and place with a desired pose for the object), \textit{path-constrained teleoperation}~(e.g., turning a valve where the path to follow is constrained based on the diameter of the valve), and \textit{trajectory-constrained teleoperation}~(e.g., arc welding where the operator should follow a velocity profile in addition to the welding path). Figure~\ref{fig:p-HRI_taxonomy} visualizes the different types of p-HRI applications.

A common and practical solution to improve workplace ergonomics is to perform risk assessments and analyze human comfort during the task. Ergonomists usually perform this process for the worst posture during the task by taking measurements of human posture onsite or from recorded images or videos. They normally use risk assessment tools~\cite{ramaganesh2021ergonomics} to assess the task and provide intervention and improvement in forms of user training~\cite{bazazan2019effect}, workstation modification~\cite{shikdar2007smart} or task rotation~\cite{motabar2021effect}. However, in most cases, it is not possible to have an ergonomist present during the entire task to perform a comprehensive risk assessment, which results in missing events or only partial understanding of awkward postures. In addition, these sampling strategies lead to failures in detecting the most hazardous combinations of force and posture, increasing the inaccuracy of an assessment. Finally, although postural correction by ergonomists generally helps reduce the risk of injuries, in many cases, they are not optimal or tailored to specific task requirements.

Early improvements on continuous monitoring of posture, such as using motion capture~(MoCap) systems to estimate the posture and task parameters~(e.g., frequency and duration of the task), still require significant time and effort for set up~\cite{alvarez2017simultaneous}. Moreover, putting markers on human operators can be inconvenient. Alternative markerless techniques are more adaptable, minimally intrusive, and less expensive. However, they need calibration to deal with errors and uncertainties from the sensors~\cite{xiao2018wearable,ganesh2021extrinsic}. Vision-based markerless methods can also be perturbed by the lighting, background color, and even the user's clothing~\cite{colyer2018review}. The proximity between the human and robot in p-HRI increases occlusion and reduces the accuracy in estimate posture as mentioned in~\cite{busch2017postural}.

Postural optimization in p-HRI has recently received substantial attention in research~(see Table~\ref{table:literature}) and developing a  model of human comfort lies at the heart of effective postural optimization. Researchers either use risk assessment tools from ergonomics or propose computational models for human comfort and ergonomics. Among all risk assessment tools, RULA~\cite{McAtamney1993rula} and REBA~\cite{hignett2000rapid} depend most on human posture and provide quantitative scores, making them good choices for postural optimization applications. However, the discrete scores and the presence of plateaus in RULA and REBA~\cite{busch2017postural} create challenges when using them in gradient-based postural optimization. Based on our experience, using the risk assessment models directly in gradient-free optimization is time-expensive, and the plateaus often prevent progress toward the optimal global solution in postural optimization. Thus, researchers in p-HRI often use computational models based on approximations of ergonomic assessment models in gradient-based postural optimizations; quadratic approximations~\cite{rahal2020caring,busch2017postural,busch2018planning} being the standard approach in the literature. However, these approximations deviate far from the scientifically validated assessments, causing doubt that they can reliably provide the same level of ergonomic benefit.

Assuming perfect ergonomics assessment and optimization, the postural correction process is also a challenging part in improving ergonomics in p-HRI. Manual correction by ergonomists is tedious and cannot be applied continuously for the human operator. More systematic approaches such as continuous visual~\cite{lorenzini2018real} and vibrotactile~\cite{kim2018ergotac} feedback to the user perform well in correcting the posture. However, they need different devices to provide feedback, require the user to accept and apply the correction. Guiding the human toward the optimal posture using the force from the robot~\cite{rahal2020caring} also may cause extra force on the human and increase the risk of injuries, as well as affecting the realistic haptic feeling.

To overcome these issues in postural estimation and optimization in p-HRI, we propose ergonomically intelligent physical human-robot interaction~(see Fig.~\ref{fig:framework}). It includes four main parts, targeting the drawbacks of current methods:
\begin{itemize}
    \item \textit{Posture estimation in p-HRI solely from the trajectory of the interacting robot}: We show that the interacting robot is an adequate sensor for continuous posture estimation of the human and for monitoring the risk of injuries. We model the posture estimation problem as a partially observable dynamical system where the robot's trajectory at the interaction point is the only observation. We provide an approach for estimating human segment lengths using interactive motion routines and a particle filter to infer the human joint angles.  
    \item \textit{Differentiable ergonomics assessment model}: We learn an accurate, continuous, and differentiable model called \textit{DULA}~(Differentiable Upper Limb Assessment) for upper body ergonomics using a standard neural network from a dataset of exhaustively sampled postures, labeled by the standard RULA risk assessment tool. 
    \item \textit{Gradient-based postural optimization}: We use DULA in a gradient-based optimization and compare it with a gradient-free optimization directly using RULA.
    \item \textit{Automatic postural correction}: We provide a new approach to correct human posture in teleoperation through the velocity scale between the leader and follower robots.
\end{itemize}

\begin{table*}[t!]
\small
\centering
\begin{tabular}{|p{0.3cm}|p{2cm}|p{3.1cm}|l|p{3.1cm}|l|}
\hline
\multicolumn{2}{|c|}{Ergonomics Model}  & \centering Analytical Models  & \centering Learned Models & \multicolumn{2}{c|}{Risk Assessment Tools}\\ \cline{1-6}
\multicolumn{2}{|l|}{Approximation} & & & \centering{Quadratic} & $\qquad$No-Approximation \\ \cline{1-6} \cline{1-6}
\multirow{6}{*}[-0.4cm]{\rotatebox[origin=c]{90}{Optimization}}&\multirow{5}{*}[-0.3cm]{\centering Gradient Based}&\textcolor{grey3}{$\medblacksquare$}\cite{peternel2018robot}&  &$\meddiamond$\cite{rahal2020caring}&\\ 
                              &&\textcolor{grey2}{$\medblacksquare$}$\medblacksquare$\cite{peternel2017towards}&This work&\textcolor{grey2}{$\medblacksquare$}\cite{busch2017postural}&\\  &&$\medsquare$\cite{chen2018planning}&&\textcolor{grey2}{$\medblacksquare$}\cite{busch2018planning}&\\
                              &&$\medblackdiamond$\cite{peternel2020human}& & &\\
                              &  &$\medblacksquare$\cite{kim2017anticipatory}&  & &\\
                              \cline{2-6}& \centering Gradient Free &&$\medsquare$\cite{marin2018optimizing}&&\textcolor{grey3}{$\medblacksquare$}\cite{van2020predicting}\\\hline
\multicolumn{2}{|c|}{No Optimization} &                                   &                    &                &$\medsquare$\cite{shafti2019real}\\ \hline                              
\multirow{2}{*}[-0.05cm]{\rotatebox[origin=c]{90}{Legend}} & \multicolumn{5}{l|}{$\medsquare$ Physical-HRI: Assistive Holding, \textcolor{grey2}{$\medblacksquare$} Physical-HRI: Handover, \textcolor{grey3}{$\medblacksquare$} Physical-HRI: Co-Manipulation} \\
&\multicolumn{5}{l|}{{$\medblackdiamond$} Teleop: Goal-Constrained with Postural Optimization By Interface Reconfiguration} \\
&\multicolumn{5}{l|}{$\meddiamond$ Teleop: Goal-Constrained with Online Postural Optimization} \\
\hline
\end{tabular}
\caption{\small State of the art of postural optimization in p-HRI and teleoperation.}
\label{table:literature}
\end{table*}
In this paper, we focus on the first three parts of our framework. We introduce our approach for general p-HRI scenarios. However, we implement and evaluate them in a simple teleoperation as an example. We initially proposed the idea of ergonomically intelligent teleoperation systems in the HRI Pioneers workshop~\cite{yazdani2021posture}. Our posture estimation for teleoperation approach was previously published at IEEE CASE conference~\cite{yazdani2020leader}, and we introduced the differentiable ergonomics model for postural optimization in the RSS Robotics for People workshop~\cite{yazdani2021dula}. In this paper, we do not add any novel experimental evolution. Instead, we formalize the framework for general p-HRI and more tightly link the previous contributions. We believe this paper brings all the previous work together, improving clarity and advancing the discussion of ergonomics in p-HRI.

\section{Related Work}
In this section we provide the state of the art in posture estimation, risk assessment and postural optimization in p-HRI.

\subsection{Posture Estimation in p-HRI}
For many years, MoCap systems using reflective markers~(\cite{das2011quantitative}) were the dominant approach for posture estimation. However, advances in computer vision have led to increased popularity in markerless techniques~\cite{saremi2018enhanced, mathis2018deeplabcut}. In p-HRI and teleoperation, researchers mainly have used external sensors~(i.e., a vision system or IMU) to estimate a user's posture, especially for hand gestures~\cite{vartholomeos2016design}.

The idea of solely using the interacting robot's trajectory~(leader robot in teleoperation) for posture estimation in p-HRI has been introduced concurrently with this research by~\cite{rahal2020caring}, where they solved the inverse kinematics of the human arm. Unlike our probabilistic approach, their heuristic does not always hold across different tasks. In comparison, our work provides a probabilistic distribution over the human posture and ergonomic risk score. Moreover, our human model covers the entire upper body, including the arm and torso. Later, \cite{vianello2021human} proposed probabilistic prediction of human postures for a given robot trajectory executed in a collaborative scenario. They learned the distribution of the null space of the human arm's Jacobian and the weights of the weighted pseudo-inverse from demonstrations. Their approach depends highly on the task and demonstrations. In comparison, our approach is task-independent and does not require any demonstration.

Defining a model for human joint limits is challenging. Studies show that the range of motion for a joint varies depending on the positions of other joints~(inter-joint dependency) or other degrees-of-freedom in the same joint~(intra-joint dependency)~\cite{wang1998three,jiang2018data} and vary by gender and person. \cite{akhter2015pose} used a dataset of recorded MoCap of human motion to develop a discontinuous mathematical model for a posture-dependant range of motion and check the validity of a full-body posture. \cite{jiang2018data} used the above model to label the validity of a set of randomly generated postures and learned a differentiable neural network based on the generated data, and used it as a constraint in the inverse kinematics optimization. Their arm model only includes the shoulder and elbow and not the wrist. We use this learned network for checking the validity of the arm posture.

\subsection{Ergonomics Models and Postural Optimizations in p-HRI}
The literature in postural optimization in p-HRI and teleoperation is minimal. Table~\ref{table:literature} summarizes the relevant literature for in both areas.

Researchers have examined ergonomics and postural optimization in three different types of p-HRI: (1) \textit{Assistive Holding} (e.g., \cite{marin2018optimizing,shafti2019real,chen2018planning}), (2) \textit{Object handover} (e.g., \cite{busch2017postural,busch2018planning,peternel2017towards}), and (3) \textit{Co-Manipulation} (e.g., \cite{van2020predicting,peternel2018robot,peternel2017towards,kim2017anticipatory}). In the first two tasks the postural optimization provides optimized posture for the human and joint configurations for the robot while maintaining contact with the object at the interaction interface. In co-manipulation, the postural optimization outputs a trajectory of optimal postures for the human and an optimal joint-space trajectory for the robot to perform the task. In teleoperation, postural optimization has only been developed for goal-constrained tasks.

Ergonomic assessment models provide the primary cost in postural optimization objectives. p-HRI researchers have proposed many computational models to assess ergonomics and comfort of users, including peripersonal space~\cite{chen2018planning}, muscle fatigue~\cite{peternel2017towards}, and joint overloading~\cite{kim2017anticipatory}. To make the optimization simpler, \cite{marin2018optimizing} suggested a contextual ergonomics model, which is a set of Gaussian process models including joint angles, moments, reaction, load, and muscle activation trained with the musculoskeletal simulation task contexts. Using Gaussian process models enables a 2D latent space to search while their cost function is defined in the high-dimensional musculoskeletal space. 

Some works use approximations of the ergonomics risk assessment tools. \cite{busch2018planning} proposed a differentiable surrogate of the REBA score by fitting a weighted combination of quadratic functions plus a constant for the task payload. \cite{rahal2020caring} suggested a quadratic approximation for RULA, which is the summation of the quadratic norm of the deviation from the human neutral posture for shoulder, elbow, and wrist joints angles. Their approximation conceptually agrees with the qualitative idea behind RULA, in which the risk score goes higher when the human deviates more from the neutral posture. 

\cite{van2020predicting} provided the only study of directly using risk assessment tools in derivative-free postural optimization. They added the REBA score to the transition cost function in an A* optimization for task and motion planning of a robot in co-manipulation. Finally, \cite{shafti2019real} identified six main causes of low ergonomic states, leading to six universal robot responses to allow the human to return to an optimal ergonomics state. Their framework identifies the causes and controls the robot to constantly adapt to the environment in an ergonomic way.

The table shows that most of the publications use analytical models and quadratic approximation to risk assessment tools in their approaches. Due to the differentiability of the models, they can be used in gradient-based postural optimization, which is faster and easier to implement for postural optimization problems. However, the reliability and optimality of the level of ergonomic benefit they provide are questionable comparing to standard risk assessment tools in ergonomics. Table~\ref{table:literature} also shows an excellent opportunity for researchers to develop accurate ergonomics models learned from standard risk assessment tools and leverage them in gradient-based optimization. We target this area of research and propose DULA, a differentiable and continuous ergonomics model that is learned from the standard assessment tool, RULA.
\section{Problem Statement}
In this section, we introduce the problem of posture estimation and optimization in p-HRI. Among the different p-HRI applications, we focus on co-manipulation and teleoperation since they have the most range of continuous motions. However, the same problem statement can be extended to other applications with minor modifications.
\subsection{Posture Estimation via the Robot Proprioceptive Sensing in p-HRI}
In p-HRI, the physical interaction occurs at the \textit{interaction point}. This is the contact point either between the human body and the robot in direct p-HRI and teleoperation or between the human body and the shared object in co-manipulation, assistive holding, and handover. In the latter case, we assume the robot firmly grasps the shared object, adding it to the robot's kinematic chain.  

We seek to solve the problem of estimating the human joint-space trajectory $\boldsymbol{\tau}=[\mathbf{q}_t;\mathbf{\dot{q}}_t]_{t=1:T}$ using only the observed task-space poses and velocities of the interaction point $\mathcal{Z}=[\mathbf{z}_t;\mathbf{\dot{z}}_t]_{t=1:T}$ from the robot. The mapping from human joint angles into the pose of the interaction point is through the kinematics of the human model parameterized by the segment length  $\mathbf{\psi}$. We estimate $\mathbf{\psi}$ independently, prior to posture estimation. Hence,
\begin{equation}
    [\mathbf{z}_t;\mathbf{\dot{z}}_t]=h(\phi([\mathbf{q}_t;\mathbf{\dot{q}}_t], \mathbf{\psi}))
    \label{eq:observation}
\end{equation}
where $h$ is the observation function, and $\phi$ is the forward kinematics of the human model. This defines only a partial observation of the human posture because of redundancy in the human kinematics and a noisy measurement at the interaction point, which may change slightly during a task.

We can write the postural estimation as an optimization:
\begin{align}
\mathbf{\tau}^*=&\argmin_{\mathbf{\tau}}\sum_{t=1}^T\!\begin{aligned}[t]
&||\phi([\mathbf{q}_t;\mathbf{\dot{q}}_t], \mathbf{\psi})-[\mathbf{z}_t;\mathbf{\dot{z}}_t]||_{\Sigma_1}^2+\\&||\left[\mathbf{q}_{t},\mathbf{\dot{q}}_{t}\right] - f(\left[\mathbf{q}_{t-1},\mathbf{\dot{q}}_{t-1}\right])||_{\Sigma_2}^2 \label{eq:pose_estimate_problem_statement}\end{aligned} \\
                & \text{s.t.} \qquad \mathbf{q}_{\mathrm{min}} \leq \mathbf{q} \leq \mathbf{q}_{\mathrm{max}} \nonumber
\end{align}
where $\mathbf{q}_{\mathrm{min}}$ and $\mathbf{q}_{\mathrm{max}}$ are the joint limits, $\Sigma$ is the weight vector for position and orientation elements, and $f$ is the human motion model defined by:
\begin{equation}
    \begin{bmatrix}
    \mathbf{q}_t\\
    \mathbf{\dot{q}}_t
    \end{bmatrix} =\begin{bmatrix}
    1&\Delta t\\
    0&1
    \end{bmatrix} \begin{bmatrix}
    \mathbf{q}_{t-1}\\
    \mathbf{\dot{q}}_{t-1}
    \end{bmatrix}+\begin{bmatrix}
    0\\
    \mathbf{\ddot{q}}_{t-1}\Delta t 
    \end{bmatrix}
\label{eq:kinematic_update}    
\end{equation}

The high degree-of-freedom in human kinematics makes this problem \emph{redundant} with an infinite number of solutions. We seek the solution closest to the actual human's posture.

\subsection{Postural Optimization in Co-Manipulation}
In co-manipulation, the robot helps a human to move an object~(usually a heavy one) from an initial pose to the desired pose~\cite{peternel2017towards}. There is no specific trajectory for the object to follow; the human plans and leads the motion, and the robot follows and provides help carrying the object. Meanwhile, the ergonomically intelligent p-HRI seeks to find the optimal posture for the human at each instance while holding the pose and velocities at the interaction point close to the current state. Once found, the system will suggest the solution to the human. We call this \textit{online postural optimization}. Using the same notation of our posture estimation approach, we can formalize finding the ergonomically optimal posture with the following optimization problem:
\begin{flalign}
\mathbf{q}^*_t = &\argmax_{\qquad\mathbf{q}_t}\quad
\mathrm{Ergonomics}(\mathbf{q}_t) \label{eq:co-manipulate_opt_online}\\
 &\text{s.t.}\quad||\phi([\mathbf{q}_t;\mathbf{\dot{q}}_t], \mathbf{\psi})-[\mathbf{z}_t;\mathbf{\dot{z}}_t]||_{\Sigma_1}^2<\epsilon \nonumber\\
            & \qquad \mathbf{q}_t \in \text{Range of Motion} \nonumber 
\end{flalign}
where ${\mathbf{q}}^*_t$ is the optimal posture at time $t$. It is important to note that the optimal posture
${\mathbf{q}}^*_t$ from Eq.~\eqref{eq:co-manipulate_opt_online} for each time step is then suggested to the human to move towards. The human can refuse or accept and try to apply the correction as much as possible while completing the task.

\subsection{Postural optimization in Teleoperation}
The main distinction of teleoperation from other p-HRI applications is the motion~(and force) coupling between the leader and follower robots. The motion coupling between the two robots is not necessarily a one-to-one scale, and the coupling can be paused and resumed. This enables the teleoperator to disengage the leader robot in the middle of a task in many types of application, reposition their postures~(and the leader robot as the result), and resume the teleoperation from a a more comfortable posture~\cite{peternel2020human}. The relative position trajectory of the follower robot remains unchanged; however, its velocity trajectory changes due to the pause. The motion scaling also helps keep the teleoperator's posture in a smaller comfort zone while the follower robot operates in a much wider zone.

To use this feature of teleoperation in postural optimization, we define three types of postural optimization, customized for different types of teleoperation from Fig.~\ref{fig:p-HRI_taxonomy}.

\subsubsection{Online Postural Optimization}
Online postural optimization in teleoperation is very similar to online postural optimization in co-manipulation. The difference is that instead of having initial and goal poses for the shared object in co-manipulation, in teleoperation, we have initial and goal poses for the end-effector of the follower robot~(or the object it manipulates). Hence, the ergonomically intelligent system should find the optimal posture of the human at each instance, in which the pose and velocities of the follower robot's end-effector are not far from the current states, and suggest it to the human. The formulation in Eq.~\ref{eq:co-manipulate_opt_online} works here too, where $[\mathbf{z}_t;\mathbf{\dot{z}}_t]$ are observed from the leader robot.

\subsubsection{Initial Postural optimization}
The remote connection between the leader and follower robots makes it possible to start the teleoperation task from any initial human posture and the corresponding leader robot's initial posture and perform the same task with the follower robot, considering the human and leader robot's workspace. We use this feature to propose \textit{initial postural optimization} for path-constrained and trajectory-constrained teleoperation tasks. In this type of postural optimization, the ergonomically intelligent system observes the human doing the task once, then calculates the optimal initial posture for the human to start \textit{the same task} with the same task-space motion for the follower robot. It also requires recording the entire motion of the robots and the human during the task.

We form initial postural optimization as the following:
\begin{flalign}
\mathbf{q}^{h^*}_0 = &\underset{\qquad\mathbf{q}^h_0}{\argmax}\quad
\sum_{t=0}^{T} \mathrm{Ergonomics}(\mathbf{q}^h_t) \label{eq:teleop_opt_initial}\\ \nonumber
&\text{s.t.} \quad ||\widehat{\dot{\mathbf{x}}}^h_t - J^h(\mathbf{q}^h_t,\mathbf{\psi}) \mathbf{\dot{q}}_t ||_{\Sigma}^2<\epsilon \quad \forall t\geq 0\nonumber
\end{flalign}
where $\widehat{\dot{\mathbf{x}}}^h_t$ is the recorded velocity of the interaction point during the performed task, and $J^h(\mathbf{q}^h_t,\mathbf{\psi})$ is the Jacobian of the human kinematic chain. Here, the superscripts $h,l,f$ refer to human, leader robot, and follower robot, respectively. 

\subsubsection{Postural Optimization by Interface Reconfiguration}
An inherent feature of path-constrained teleoperation is the ability to pause the teleoperation by disengaging the leader and follower robots from each other, move the leader robot to a new position, and resume the teleoperation. This is usually done by a clutch switch placed under the user's foot. Benefiting from this feature, we propose \textit{postural optimization by interface reconfiguration}, in which, when the ergonomically intelligent system detects high-risk postures during a previously recorded task, suggests to pause the teleoperation. When teleoperation is paused, the system calculates a new optimal posture for the human to resume the teleoperation from it and continue the task. 

The formulation is similar to the initial postural optimization but covers the time after the pause:
\begin{flalign}
\mathbf{q}^{h^*}_{t_p+1} = &\underset{\qquad\mathbf{q}^h_{t_p+1}}{\argmax}\quad
\sum_{t=t_p+1}^{T} \mathrm{Ergonomics}(\mathbf{q}^h_t) \label{eq:teleop_opt_reconfig}\\ \nonumber
&\text{s.t.} \quad ||\widehat{\dot{\mathbf{x}}}^h_t - J^h(\mathbf{q}^h_t,\mathbf{\psi}) \mathbf{\dot{q}}_t ||_{\Sigma}^2<\epsilon \quad \forall t>t_p \nonumber
\end{flalign}
where $t_p$ is the time at pause in teleoperation.
\section{Approach and Implementation}
\label{sec:approach}
This section proposes our approach to solving the problems mentioned in the previous section and details their implementations.

\subsection{Human Kinematics Model}
\label{sec:approach-kinematic}
We use a 10-DOF kinematics model~(see Fig.~\ref{fig:network}) to analyze the upper-body motion of a human sitting on a chair and interacting with a robot. We assume that the chair is stationary with a known position w.r.t. the robot, and the human sits on the center of the chair. In~\cite{yazdani2020leader}, we proposed an approach based on circle point analysis~\cite{mooring1991fundamentals} for estimating human segment lengths in p-HRI using interactive motion routines and we use it here. We encode human motion limits based on both fixed limits and the posture-dependant joint limit model from \cite{jiang2018data}.


\subsection{Particle Filter for Posture Estimation}
\label{sec:approach-pf}
We approximate the solution for the partially-observable problem of posture estimation in Eq.~\ref{eq:pose_estimate_problem_statement} by a particle filter ~\cite{thrun2005probabilistic}. As the estimation of the human model from the trajectory of the leader robot has high ambiguity due to the redundancy, we add the joint angular velocities to our state variables and use the velocity of the robot's end-effector as a part of the observation. As a prior, we encode that the human \textit{starts} the task in a static and neutral posture. We initialize $M$ particles using a truncated normal distribution with the mean at the neutral posture $\mathbf{q}_{neutral}$ and set the initial angular velocities to zero.

\subsection{DULA: Differentiable Upper Limb Assessment Model for Postural Optimization}
Solving Eqs.~\ref{eq:co-manipulate_opt_online}, \ref{eq:teleop_opt_initial}, \ref{eq:teleop_opt_reconfig} using a gradient-based optimization requires a differentiable and continuous ergonomics model as objective functions. Instead of using analytical models or approximations, we introduce \textit{DULA}~(Differentiable Upper Limb Assessment) which is an accurate, continuous, and differentiable model for upper body ergonomics. It is learned using a standard neural network from a dataset of exhaustively sampled postures, labeled by RULA. 

To learn a continuous and differentiable model for RULA, we designed a fully connected regression-based neural network. While RULA provides discrete integer scores from 1 to 7, we choose to predict continuous labels. If we had instead performed multi-class classification based on the discrete labels, the resulting model would be less useful in optimization as there is no natural choice of a smooth objective to minimize or constrain ergonomic cost. This would negate our desire for a computationally helpful model. Hence, we perform regression to the multi-class labels.  
\begin{figure}[t]
  \includegraphics[width=8.4cm]{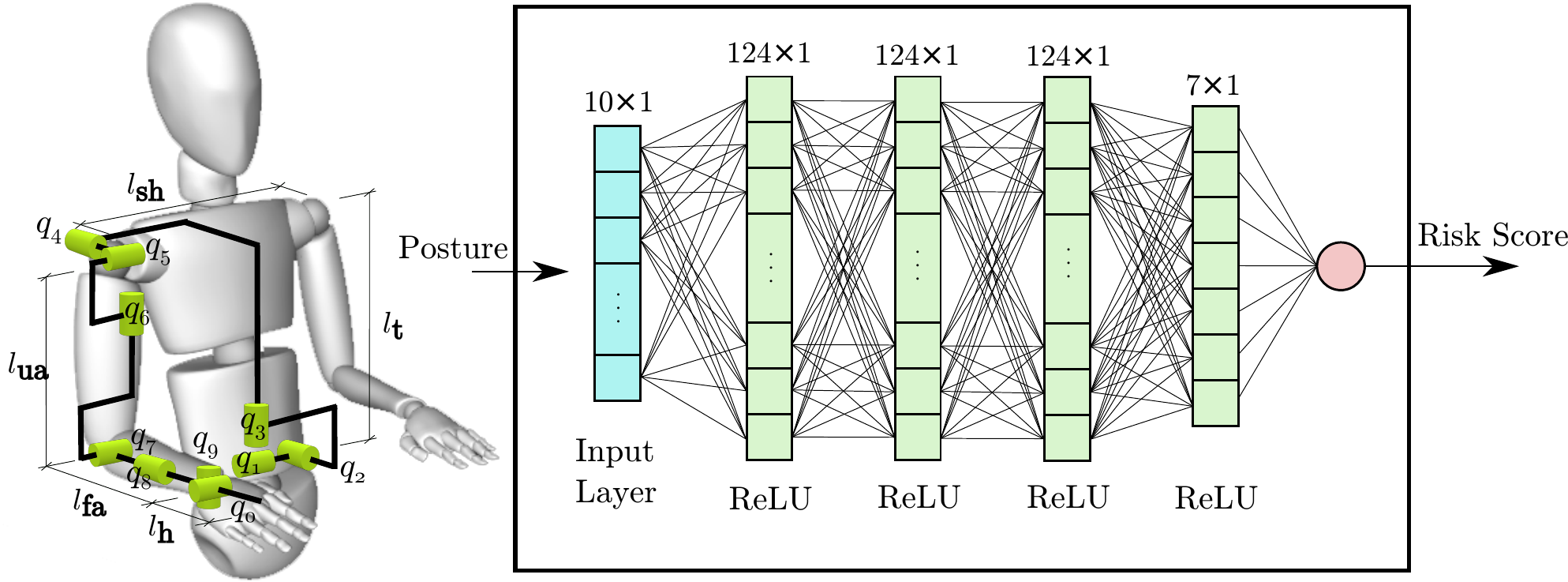}
  \caption{\small The structure of the DULA neural network.}
  \label{fig:network}
\end{figure}

The structure of the neural network is shown in Fig.~\ref{fig:network}. It includes 4 hidden layers with ReLU activation function. We found that a network with 124 units for the first three layers and 7 units for the last hidden layer worked best. We train the network using the standard mean squared error loss function for 2000 epochs using a learning rate of 0.001. We used 5-fold cross-validation to find the optimal network parameters.

We developed a dataset of 7.5 million upper body postures of a human. We additionally defined task parameters based on the RULA worksheet---the frequency of the arm and body motions; type and maximum load on the arm and body; neck angle; and whether any legs, feet, or arms are supported. We developed a script for automatic RULA assessment based on the posture and tasks parameters and verified it with several ergonomists. We used this script to label the posture dataset. Since postures with labels 1, 2, 6, and 7 are not frequent in the full range of human motion, we balanced the dataset by forcing the data generation scripts to generate enough data points with those labels. We split the dataset into 80\% training and 20\% testing sets. 

Training of the neural networks takes about 60 hours on a Core i9 CPU with GPU support. We implemented and trained our network using PyTorch~\cite{NEURIPS2019_9015}.

\subsection{Postural optimization}
We explore two methods for postural optimization. Initially, we use the cross-entropy method~(CEM) \cite{kobilarov2012cross, de2005tutorial} as a sample-based and gradient-free optimization. In our implementation, we use a Gaussian probability distribution N=10000 samples.

As an alternative method, we utilize a derivative-based optimization using our learned DULA model. We solve the nonlinear and constrained least-square postural optimization using sequential least squares programming~(SLSQP)~\cite{nocedal2006numerical} solver in SciPy. We calculated DULA gradients using automatic differentiation in PyTorch.

The following section examines variations of postural correction optimizations using these two methods and compares the resulting RULA score distributions.
\section{Evaluation \& Experimental Protocol}
We evaluated our proposed postural estimation and optimization approaches through human subject studies and simulation experiments, respectively.
\label{sec:implementation}
\subsection{Human Subject Study for Posture Estimation}
\begin{figure}[t!]
\center
\includegraphics[width=8.3cm]{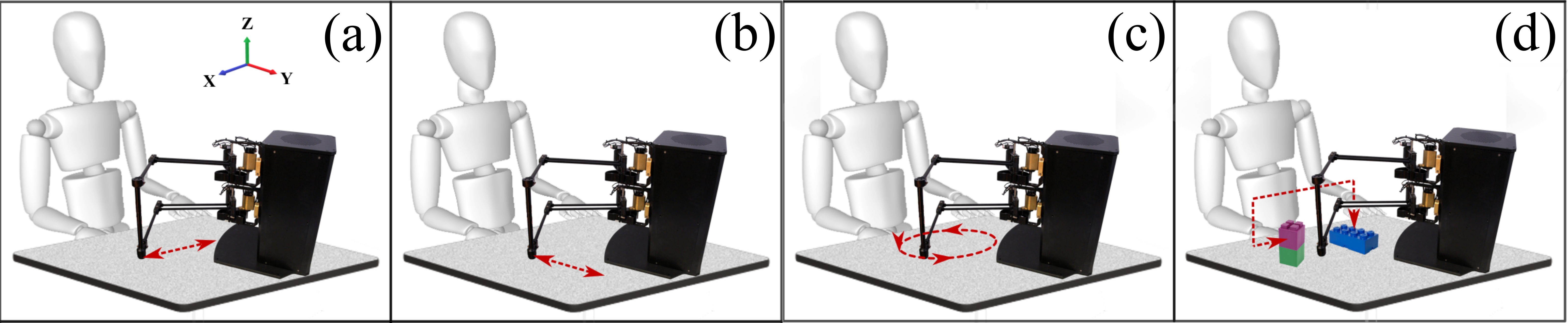}
\caption{\small The tasks for human subject study: repetitive motions following a straight line in the $X$ direction~(a), a straight line in the $Y$ direction ~(b), a circular path~(c), and repetitive motions between random sides of two blocks positioned at different heights, with an \textit{unprescribed} motion and high range of hand rotation~(d).} \label{fig:exp_tasks}
\end{figure}

\begin{figure}[t!]
\begin{centering}
\includegraphics[width=8.2cm]{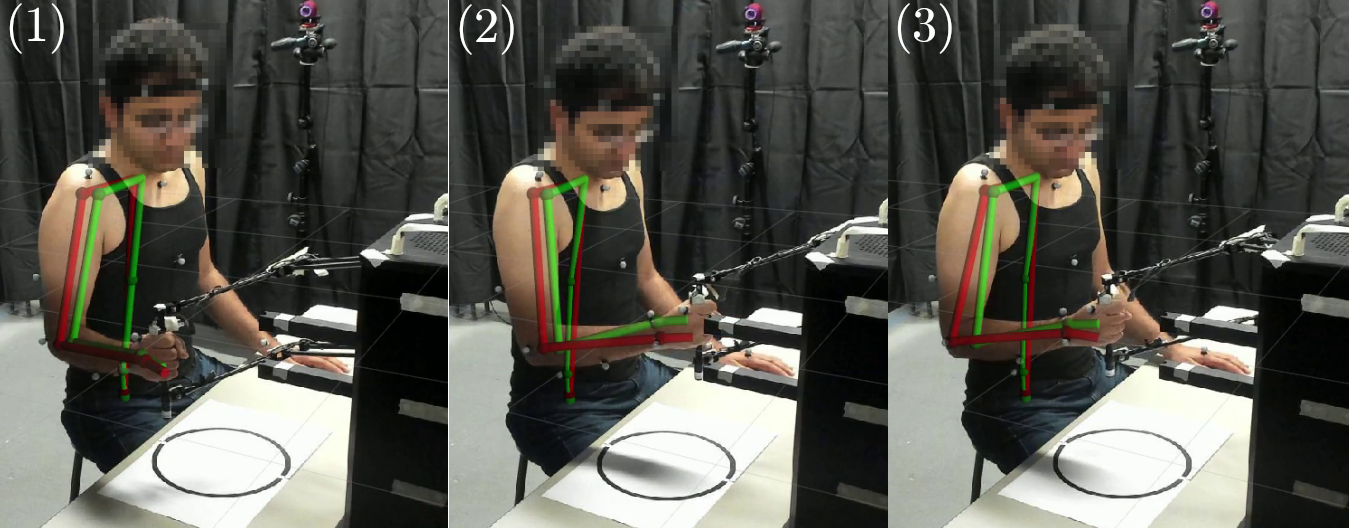}
\caption{\small Video-overlaid skeletons show the posture estimated from our proposed approach~(red) and MoCap~(green).}
\label{fig:frames}
\end{centering}
\end{figure}

We conducted a human subject experiment in which participants interact with a 6-DOF Quanser HD\(^{2}\) haptic interface as the leader robot~(see Fig.~\ref{fig:frames}) in teleoperation, and we recorded their upper body motion using a 12-camera Optitrack MoCap system for comparison. We recruited 8 participants~(4 females, 4 males) with ages ranging from 25 to 33 years and heights in the range of $171\pm21$cm. Participants were graduate students from various programs and did not have any experience with teleoperationing robots and received 15 minutes of training with the robot. Each participant performed 4 tasks visualized in Fig~\ref{fig:exp_tasks}. We provided a printed visual guide on the table for the first three tasks. However, the participants were not required to follow the path accurately. The participants were not told what posture to initialize the task from and how high they should be above the table to do the task. The robot collected data from the participant's motion.

To compare with other well-known approaches, we solved the postural estimation problem using two other deterministic methods: (1) boosted and bounded online least-squares IK optimization~(\textit{Online-IK}) in which we simply solve the inverse kinematics optimization independently at each time step by initializing it with the solution from the previous step, and (2) boosted and bounded offline least-squares trajectory IK optimization~(\textit{Offline-TrajIK}) in which we solve the inverse kinematics problem for the whole trajectory by initializing it with the solution trajectory from Online-IK. We used the dogleg algorithm with rectangular trust regions \cite{nocedal2006numerical} in SciPy as the solver.


\subsection{Simulated Environment For Postural Optimization}
We developed an open-source simulator for postural correction using ROS. It includes a human seated on a stool and a KUKA LWR-4 robot~(two robots for teleoperation). The simulated human behaves like a human in two ways: (1) physically controlling the interactive task and (2)~accepting or rejecting the recommended optimal postural corrections. 

\begin{figure}[t]
\center
\includegraphics[width = 8.1cm]{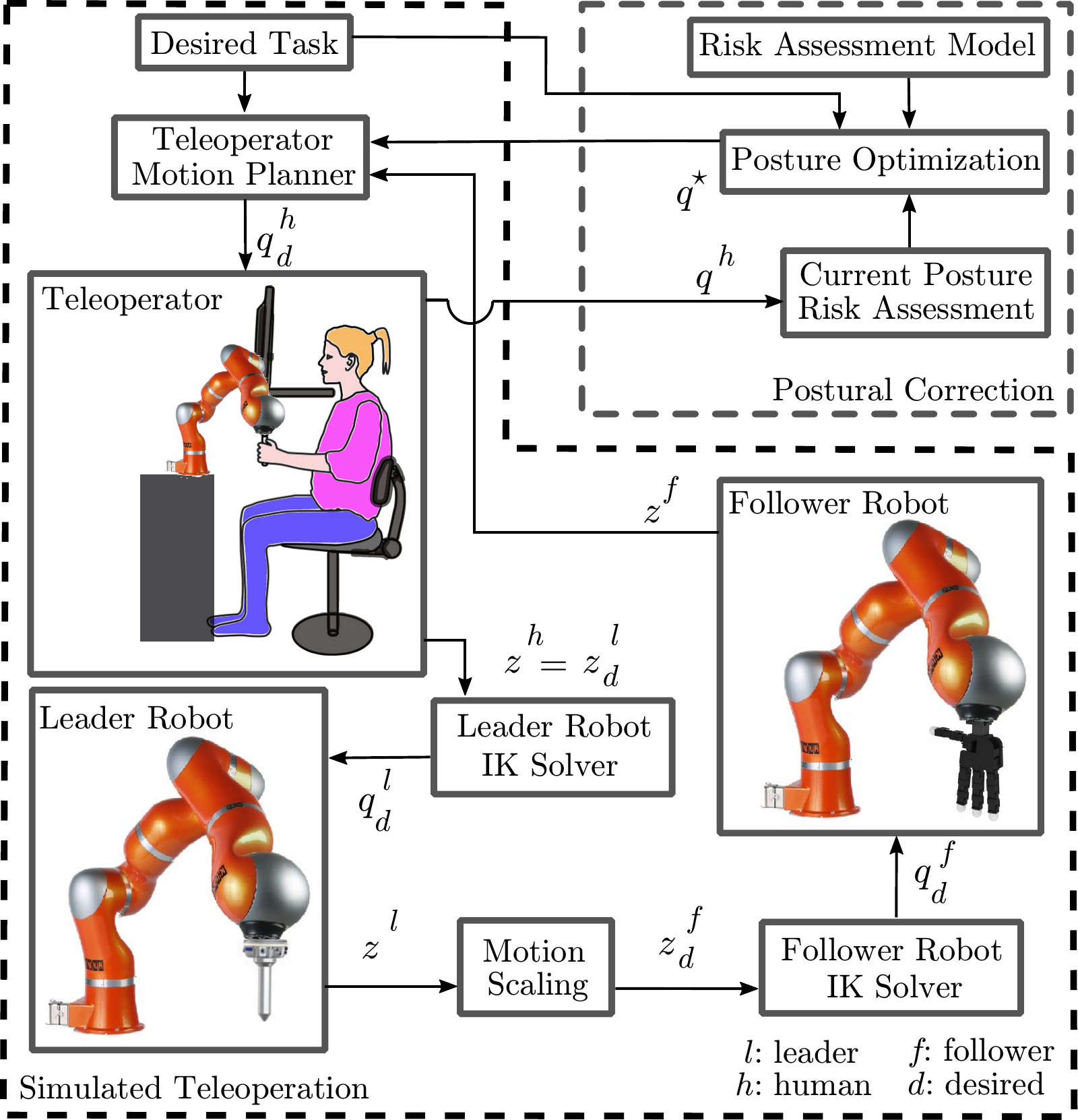}
\caption{\small Overview of teleoperation simulation.}
\label{fig:teleop-framework}
\end{figure}

We model this in an optimal motion planning framework with re-planning that finds a human joint trajectory that interacts with the robot to do the task while moving toward the optimal ergonomic posture. For example, it is formulated as the following for a teleoperation task:
\begin{flalign}
\mathbf{\tau}^{h^*}_{t\rightarrow H} = \text{arg}\underset{\mathbf{\tau}^h_{t\rightarrow H}}{\text{min}}\sum_{t=t}^H||\mathbf{x}^f_g - \mathbf{x}^f_t||^2_{\Sigma}  + \alpha||\mathbf{q}^h_t - \mathbf{q}^{h^*}_t||_2^2
\end{flalign}
where $\mathbf{\tau}^f_{t\rightarrow H}$ is the trajectory of human posture from time $t$ to the time of the horizon $H$, $\mathbf{x}^f_g$ is the goal pose of the follower robot, and $^f\mathbf{x}_t$ is the pose of the follower robot at time $t$. Here, $0\leq \alpha \leq 1 $ is a scalar number that models the postural correction acceptance and the effort of the human operator towards applying the postural correction. Details of the motion planning for the humans and the robots are presented in Fig.~\ref{fig:teleop-framework}. We note that this model is likely a simplification of true human interactive behavior. We do not advocate for its use over human subject studies. Instead, we propose it as a useful tool for systematically exploring new human safety assessment and improvement algorithms.
\begin{figure}[t]
\begin{centering}
\includegraphics[width=8.6cm]{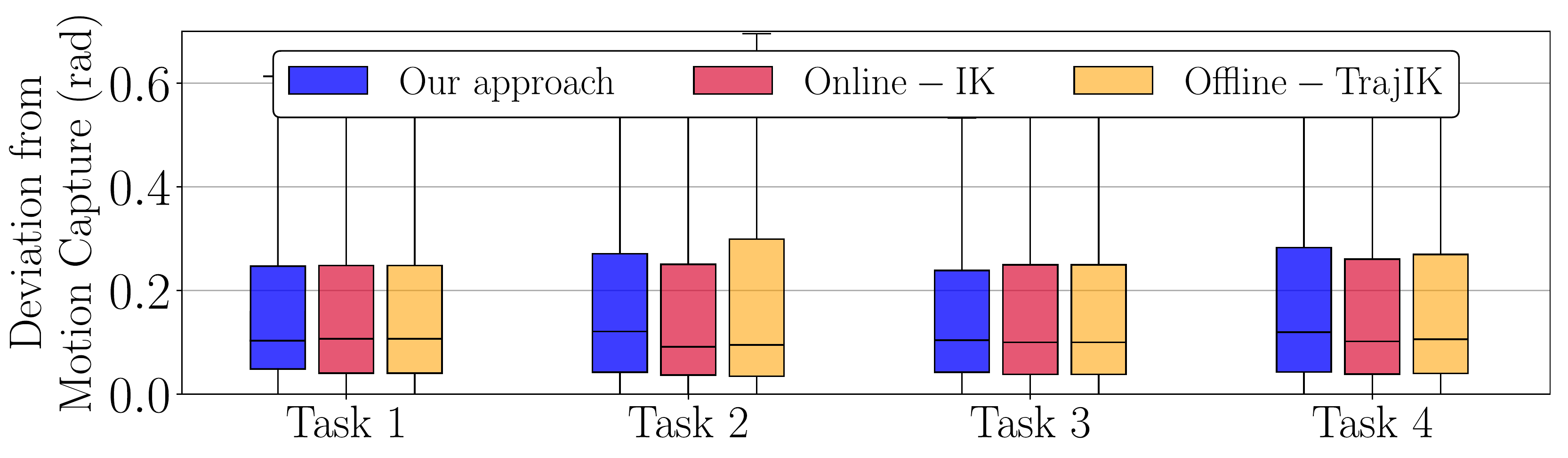}
\caption{\small Deviation of the posture estimated by our approach vs Online-IK and Offline-TrajIK methods among all the tasks.}
\label{fig:ik_methods_tasks}
\end{centering}
\end{figure}
\section{Results}
This section provides results for our ergonomically intelligent p-HRI system. We discuss the proposed posture estimation approach's performance compared with MoCap, and their corresponding estimated risk assessment results. Moreover, we showcase the accuracy of the learned model for DULA and its effect on postural optimization.
\subsection{Posture Estimation}
Fig.~\ref{fig:ik_methods_tasks} compares our posture estimation approach with two other least-squares IK solutions for redundant robots for all participants among 4 tasks. From the statistical analysis of the results, we can conclude that our probabilistic \textit{particle filter} approach performance is not significantly different from the other two methods.  Overall, the approach generally agrees with MoCap with a median deviation less than 0.09rad~(less than 5deg) and upper quartile less than 0.25rad~(less than 15deg) considering the observation solely from the stylus trajectory and having no extra sensors.

Since there is no ground truth for human posture to compare with, we provide a qualitative evaluation using video frames of a participant during task~(3) with overlaid reconstructed skeletons from MoCap~(green) and our approach~(red) in Fig.~\ref{fig:frames}. The estimated postures align well with the MoCap postures during the entire task. 

\begin{figure}[t!]
\begin{centering}
\includegraphics[width=8.3cm]{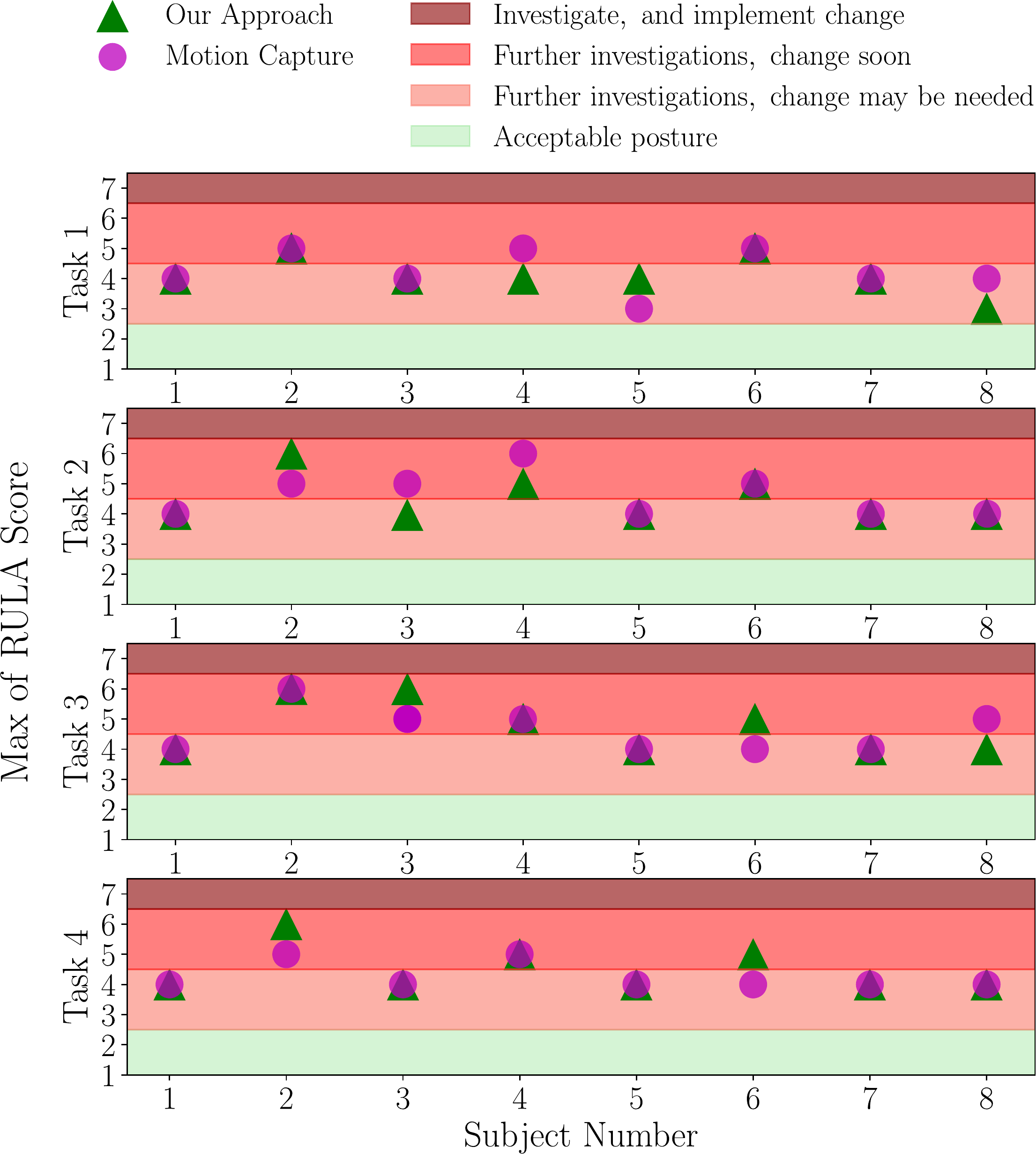}
\caption{\small Comparison of the maximum value of RULA scores of a task using estimated posture from our approach and MoCap for all the participants and trials.}
\label{fig:rula_subject}
\end{centering}
\end{figure}

Fig.~\ref{fig:rula_subject} shows the maximum RULA score during a task for the postures from our approach and MoCap postures. Our approach successfully identified all instances where the score was higher than 2~(i.e., future investigation or change may be or is needed) in all 32 trials. It also resulted in the same interpretation of the RULA in 27 trials~(84.37\%) and the same RULA score in 21 trials~(65.63\%).

Overall, the results showed that our posture estimation solely from the robot has the potential to be used for continuous monitoring of ergonomics in p-HRI and rising alerts when further investigation or change in the task is required.

\subsection{Postural Optimization}
The trained model for DULA performs well on the test dataset and results in 99.73\% accuracy. In the accuracy test, we round our continuous output to the nearest integer when reporting accuracy. The lowest diagonal element in the confusion matrix for DULA is 99.38\%, which shows the high accuracy of the learned model across all ranges.

\begin{figure}[t]
\center
\hspace{-4mm}
\includegraphics[width = 8.3cm]{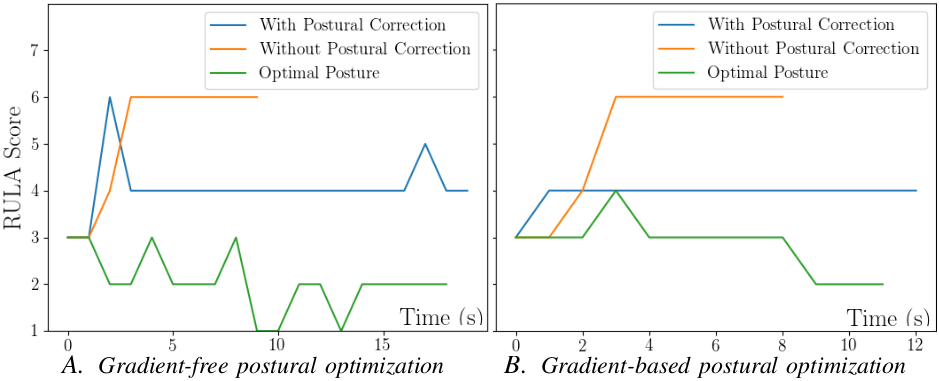}
\caption{\small Comparison between gradient-free and gradient-based postural correction on a teleoperation task. Lower scores are better.}
\label{fig:result1}
\end{figure}

Figure~\ref{fig:result1} shows the postural correction results for gradient-free~(left) and gradient-based~(right) optimization in a simple teleoperation task. It presents the RULA scores for calculated optimal posture~(green), human posture without postural correction~(orange), and the human posture after applying the correction~(blue) according to the human control and acceptance model during the task using $\alpha\sim\{0,0.75\}$. The plot shows that the risk is reduced after the simulated human applies the suggested optimal postural correction. Also, the task completion time increases without a significant decrease in risk. We can also see that the gradient-based approach provides a smoother motion concerning the RULA risk score and avoids going through postures with a risk higher than 4. It also results in a shorter task completion time.

\begin{figure}[t]
\center
\includegraphics[width = 8.2cm]{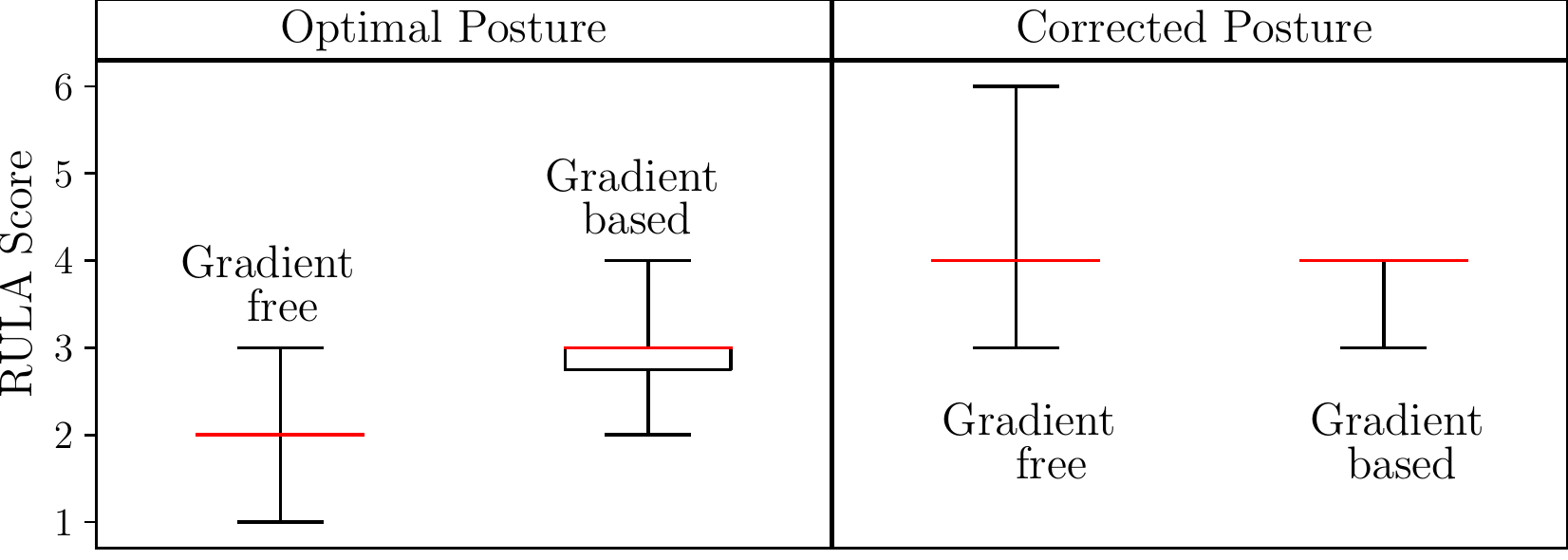}
\caption{\small Comparison in optimal posture and corrected posture between gradient-free and gradient-based optimization.}
\label{fig:comparison}
\end{figure}

Figure~\ref{fig:comparison} provides more information on comparing gradient-free and gradient-based postural optimization. It shows that the median risk scores of target optimal postures from the gradient-free approach is lower. However, the postures of the simulated human are more comfortable after applying the optimal posture calculated from the gradient-based optimization. We believe this is due to the smoother optimal postures that has been suggested to the simulated human from the gradient-based method. Moreover, the gradient-based approach is much faster. Each iteration of the gradient-free approach using 10000 samples takes 2.7 minutes to solve, while the gradient-based approach takes only tenths of a second.

\section{Conclusion}
This paper proposed a framework for ergonomically intelligent physical human-robot interaction that includes posture estimation, risk assessment, and postural optimization. We showed that the interacting robot is an adequate sensor to continuously monitor posture to assess the ergonomics of an interactive task. We described a probabilistic approach based solely on the trajectory of the robot, which is already necessary to perform the p-HRI task. Additionally, we introduced DULA, a differentiable and continuous ergonomics model to assess human upper body posture. DULA is learned to replicate the non-differentiable RULA assessment tool using a neural network, and it is 99.73\% accurate and computationally designed for efficient postural optimization for p-HRI and other related applications. We also introduced a framework for postural optimization in p-HRI using DULA and presented a demo task. The results reveal that postural optimization using DULA lowers the risk of injuries.

There are several directions for future work. For the posture estimation part, Our approach can be used for full arm representation in virtual reality systems based on the pose of controllers and headsets. When available, it would be straightforward to combine our proposed approach with other sensing modalities (e.g., vision or MoCap). We are examining the combination of our approach with OpenPose~\cite{cao2019openpose} using incremental smoothing~\cite{kaess2012isam2} in order to improve estimation results and decrease runtime. Our risk assessment work can be immediately extended by following the same procedure for REBA, a whole-body assessment tool. Additionally, we intend to conduct a human subject study to evaluate our postural optimization and correction approach. We wish to compare different means of feedback for the posture correction to the human, including visual, auditory, and haptic feedback.
\bibliography{AI-HRI2021}
\end{document}